\def\year{2022}\relax
\documentclass[letterpaper]{article} % DO NOT CHANGE THIS
\usepackage{aaai22}  % DO NOT CHANGE THIS
\usepackage{times}  % DO NOT CHANGE THIS
\usepackage{helvet}  % DO NOT CHANGE THIS
\usepackage{courier}  % DO NOT CHANGE THIS
\usepackage[hyphens]{url}  % DO NOT CHANGE THIS
\usepackage{graphicx} % DO NOT CHANGE THIS
\usepackage{amsfonts} 
\usepackage{amsmath}
\usepackage{amssymb}
\usepackage{natbib}  % DO NOT CHANGE THIS AND DO NOT ADD ANY OPTIONS TO IT
\usepackage{caption} % DO NOT CHANGE THIS AND DO NOT ADD ANY OPTIONS TO IT
\DeclareCaptionStyle{ruled}{labelfont=normalfont,labelsep=colon,strut=off} % DO NOT CHANGE THIS
\usepackage{algorithm}
\usepackage{algorithmic}
\usepackage[switch]{lineno}  % add line numbers
\usepackage[pagebackref=true,breaklinks=true,letterpaper=true,colorlinks,citecolor=blue,urlcolor=red,bookmarks=false]{hyperref}
\usepackage{newfloat}
\usepackage{listings}

\usepackage{multirow}
\usepackage{siunitx,booktabs,caption}
\usepackage{graphicx}
\usepackage{amsmath}
\usepackage{amssymb}

\usepackage{setspace}
\usepackage{ctable}
\usepackage{makecell}

\title{Learning Optical Flow with Adaptive Graph Reasoning}
\author{
	Ao Luo\textsuperscript{\rm 1}, Fan Yang\textsuperscript{\rm 2}, Kunming Luo\textsuperscript{\rm 1}, Xin Li\textsuperscript{\rm 2}, Haoqiang Fan\textsuperscript{\rm 1}, Shuaicheng Liu\textsuperscript{\rm 3,1}\thanks{Corresponding author}\\
}
\affiliations{
	\textsuperscript{\rm 1}Megvii Technology~~~~~~\textsuperscript{\rm 2}Group 42\\
	\textsuperscript{\rm 3}University of Electronic Science and Technology of China\\
}
\usepackage{bibentry}
\begin{document}

\maketitle

%\linenumbers

\begin{abstract}
	
Estimating per-pixel motion between video frames, known as optical flow, is a long-standing problem in video understanding and analysis. Most contemporary optical flow techniques largely focus on addressing the cross-image matching with feature similarity, with few methods considering how to explicitly reason over the given scene for achieving a holistic motion understanding. In this work, taking a fresh perspective, we introduce a novel graph-based approach, called adaptive graph reasoning for optical flow (AGFlow), to emphasize the value of scene/context information in optical flow. Our key idea is to decouple the context reasoning from the matching procedure, and exploit scene information to effectively assist motion estimation by \emph{learning to reason} over the adaptive graph. The proposed AGFlow can effectively exploit the context information and incorporate it within the matching procedure, producing more robust and accurate results. On both Sintel clean and final passes, our AGFlow achieves the best accuracy with EPE of 1.43 and 2.47 pixels, outperforming state-of-the-art approaches by 11.2\% and 13.6\%, respectively. Codes will be publicly available at \url{https://github.com/LA30/AGFlow}.
	
\end{abstract}

% ------------------------------
\section{Introduction}

Optical flow is a fundamental task in video understanding and analysis, aiming to estimate the pixel-wise correspondence between two video frames. It has drawn continuous attention from both academia and industry due to its wide applications, \emph{e.g.,}  person-identification~\cite{chen2020frame}, visual tracking~\cite{vihlman2020optical} and video inpainting~\cite{Xu_2019_CVPR}. Recent years have witnessed significant breakthroughs made to push its performance frontier~\cite{Dosovitskiy2015FlowNetLO, Ilg2017FlowNet2E,Teed2020RAFTRA, Jiang2021LearningOF}, but it remains challenging due to inherent ambiguity in textures, large displacements, occlusions, motion blur, and non-Lambertian effects. 

\begin{figure}[t]
	\begin{center}
		\includegraphics[width=0.98\linewidth]{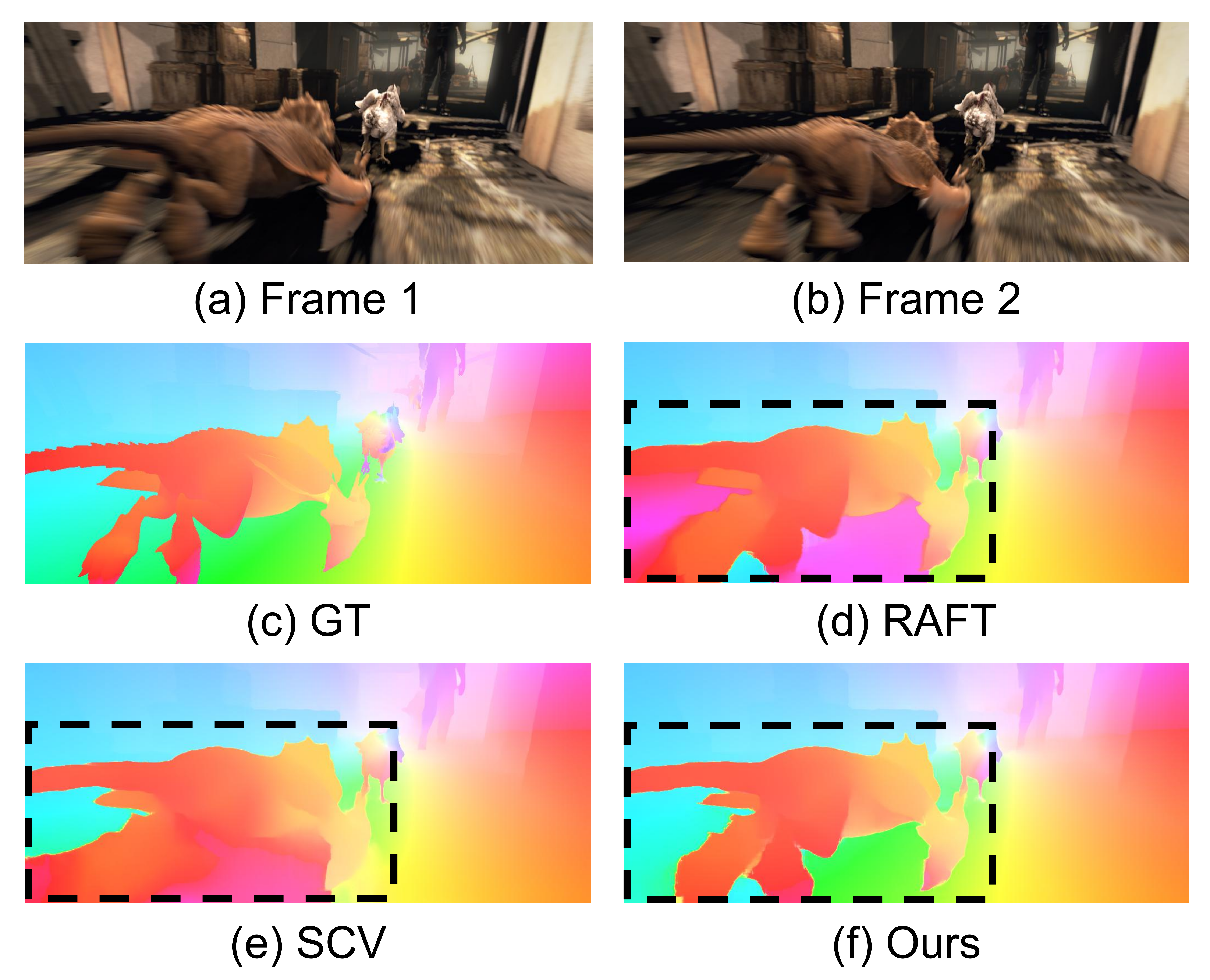}
	\end{center}
	\vspace{-.5em}
	\caption{A challenging image pair with heavy motion blur from the final pass of Sintel {\tt test} set. Unlike previous state-of-the-art methods (RAFT~\cite{Teed2020RAFTRA} and SCV~\cite{Jiang2021LearningOF}) suffering from ambiguous matching, our AGFlow is able to perform global reasoning conditioned on scene context for achieving a holistic motion understanding.}
	\label{fig:teaser}
	\vspace{-1em}
\end{figure}

Traditional optical flow algorithms formulate the dense matching as an energy minimization problem based on feature constancy and spatial smoothness~\cite{Horn1981DeterminingOF, Brox2004HighAO, Bruhn2005LucasKanadeMH}. However, because the hand-designing features and optimization objectives are difficult to cover all scenarios, these approaches are not robust enough to deal with complex motions. As a powerful alternative to traditional methods, deep learning based approaches take the research of optical flow into a new level. Current optical flow methods have made great advances in: {\bfseries i)} developing powerful data-driven, end-to-end learning paradigms~\cite{Dosovitskiy2015FlowNetLO,Ilg2017FlowNet2E};  {\bfseries ii)} designing multiple refinement strategies~\cite{Sun2018PWCNetCF,Hur2019IterativeRR,Zhao2020MaskFlownetAF}; {\bfseries iii)} exploiting auxiliary information from related tasks~\cite{zhao2020msrn}; and {\bfseries iv)} modeling pixel-wise relations for all pairs~\cite{Teed2020RAFTRA, Jiang2021LearningOF}. Although these deep learning based approaches have shown the strong capability of matching across frames, they are subject to a significant limitation: current methods largely focus on addressing the matching similarity between features, lacking a holistic motion understanding of the given scene. Thus, the ambiguities (local variations) caused by motion blur, occlusion and large motions severely degrade the accuracy of current models (see Fig.~\ref{fig:teaser}), hindering their applications in the real world. 

Taking a closer look at current optical flow methods~\cite{Sun2018PWCNetCF, Yang2019VolumetricCN}, their success is majorly attributed to an important component, namely 4D correlation volume, that typically model the correlations between features across frames. They expect the correlations achieved by using deep trainable features to surmount all ambiguities. However, evidence from~\cite{Teed2020RAFTRA} indicate that such correlations are vulnerable to variable feature representations on challenging conditions. To better aggregate context information within motion boundaries, existing arts~\cite{Teed2020RAFTRA,Jiang2021LearningOF} inject image features by concatenation operation, and encode scene information by using extra convolutional layers. But despite these attempts, the holistic motion understanding is far from being achieved: {\bfseries i)} existing approaches capture scene information with only naive operations, \emph{e.g.,} feature stacking, without typed functions for explicitly modeling such process; {\bfseries ii)} their operations are confined to the original coordinate space, causing a heavy computational burden and lacking a global understanding of the given scene; and {\bfseries iii)} they ignore the adverse effect caused by `domain gap', \emph{e.g.,} the gap between scene content and motion features. These observations prompted us to think about: \emph{how to empower the optical flow model to effectively obtain the capability of holistic motion understanding?} 

To answer this question, we propose to conduct \emph{adaptive graph reasoning} within the conventional optical flow framework. Our idea is to decouple the context reasoning from the matching procedure, and explore scene information to effectively assist in motion estimation by \emph{learning to reason} over the adaptive graph. We argue that a powerful optical flow model should have the capability of going beyond regular grids, which learns to understand the real-world motions under the guidance of scene content from a more global view. Towards this goal, we introduce a novel graph-based approach, namely adaptive graph reasoning for optical flow (AGFlow), which embeds graph techniques onto the matching pipeline to enable an effective context reasoning and information interaction. The proposed AGFlow learns to match features conditioned on scene context, and allows objects' spatial extents to be well aggregated and thus largely decreases the uncertainty of ambiguous matching.

In particular, AGFlow consists of three major components: a motion encoder that maps input frames into high-level representations for computing their 4D correlation volumes, a context encoder that extracts features only from the first input frame for capturing scene information, and an adaptive graph reasoning (AGR) module that learns to understand the given scene and distills useful information to assist in optical flow estimation. Unlike existing approaches~\cite{Teed2020RAFTRA,Jiang2021LearningOF}, our AGR exploits scene information for optical flow in the graph domain; that is, both context and motion features are mapped from regular grids to graph space for learning optical flow together. It enables the model to understand the motion from a larger context with only limited extra parameters and FLOPs (see Tab.~\ref{tab:abla}). Importantly, to overcome the domain gap between context and motion representations, a novel graph adapter (GA) is introduced to adapt (motion) graph parameters in a one-shot manner, incorporating the scene information flexibly.  Overall, our approach helps optical flow models conduct more efficient scene understanding and naturally distill scene information to assist in optical flow estimation, eventually leading to strong holistic-motion-understanding capability and better performance.  The {\bfseries contributions} of this work are summarized as follows:
\begin{itemize}
	\item[$\bullet$] 
	{\bfseries A novel graph-based approach for optical flow.} To our knowledge, this is the first work that explicitly exploits scene information to assist in optical flow estimation by using graph techniques. The proposed AGFlow can go beyond the regular grids and reason over the graph space to achieve a better motion understanding, thus successfully handling different challenges in optical flow. 
	
	\item[$\bullet$] 
	{\bfseries An adaptive cross-domain graph reasoning approach.} In order to incorporate scene information, we generalize the \emph{learning to adapt} mechanism~\cite{Bertinetto2016LearningFO} from regular grids to the graph domain. Our designed graph adapter can fast adapt scene context to guide the global (motion) graph reasoning in a one-shot manner.
	
	\item[$\bullet$] 
	{\bfseries State-of-the-art results on widely-used benchmarks.} Our AGFlow sets new records on both Sintel and KITTI benchmarks, outperforming state-of-the-art approaches by a relatively large margin.
\end{itemize}

\begin{figure*}[ht]
	\begin{center}
		\includegraphics[width=0.98\linewidth]{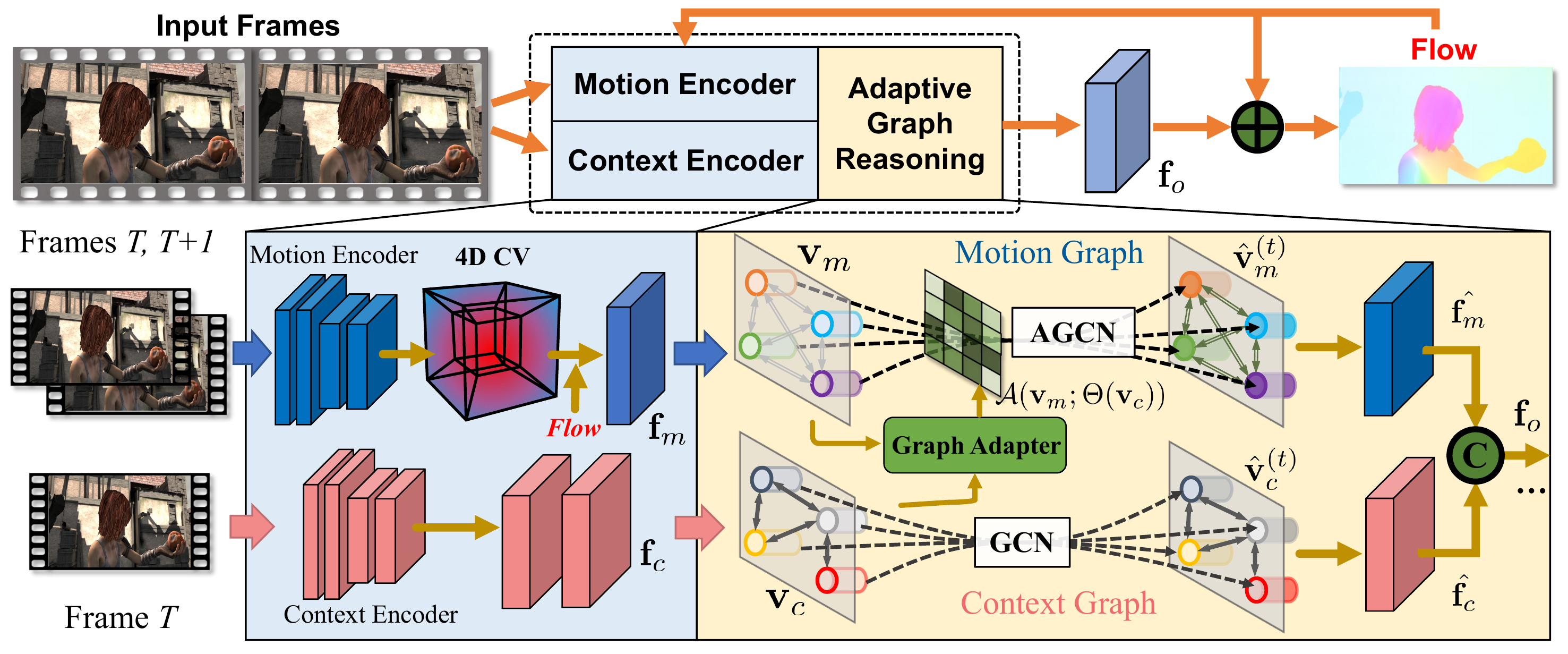}
	\end{center}
	\vspace{-1em}
	\caption{Architecture of the proposed adaptive graph reasoning for optical flow (AGFlow). ``4D CV'' means 4D correlation volumes, ``C'' indicates concatenation and ``$\bigoplus$'' denotes summation. Please refer to Sec.~{\color{red}3} for more details about the feature notations in this figure. Best viewed in color.}
	\vspace{-1em}
	\label{fig:2}
\end{figure*}

% ------------------------------
\section{Related Work}

\noindent{\bfseries \small Optical Flow Estimation.}
Optical flow is the task of estimating per-pixel motion between video frames. In the early stage, researchers~\cite{black1993framework,Horn1981DeterminingOF,Brox2004HighAO, Bruhn2005LucasKanadeMH} consider this task as an energy minimization problem, with the goal of achieving an ideal tradeoff between feature similarities and motion smoothness. However, as motion itself is hard to be modeled/described by handcrafted features and optimization objectives, it is challenging to obtain precise flow fields by traditional methods. In the deep learning era, early attempts mainly focus on {\bfseries i)} learning more robust data terms~\cite{bai2016exploiting,weinzaepfel2013deepflow} or {\bfseries ii)} avoiding the optimization step to directly estimate optical flow~\cite{Teed2020RAFTRA, Jiang2021LearningOF}. To improve results on optical flow, many recent works introduce stronger learning paradigms that can enable iterative refinement~\cite{ranjan2017optical,Sun2018PWCNetCF,Yang2019VolumetricCN,Hui2018LiteFlowNetAL}, explicit pixel-wise-relation modeling, and joint representation learning with other tasks~\cite{zhao2020msrn}. Although remarkable progress have been achieved by these developments, there is still a large room for improvement over existing approaches that largely focus on addressing the matching similarity between features without considering how to achieve a holistic motion understanding. Taking a further step, our AGFlow is empowered to exploit and incorporate high-level scene information to predict optical flow, linking the low-level matches with high-level semantic information for better accuracy. 

\noindent{\bfseries \small Graph Neural Networks.} 
Graph Neural Networks have been widely applied to different applications, including person re-identification~\cite{shen2018person}, salient object detection~\cite{luo2020cascade}, 3D shape analysis~\cite{wei2020view}, semantic segmentation~\cite{luo2021robust} and video question answering~\cite{Park_2021_CVPR}. In the context of optical flow, Graph Convolutional Networks (GCNs) are still largely under-explored. This is because most existing approaches largely focus on the matching similarity between features; the high-level scene information is overlooked or poorly investigated. In this paper, we show that GCNs can benefit the optical flow by comprehensively mining the scene information to assist in flow estimation. It helps the optical model to go beyond regular grids and understand the motion from a global view. Unlike most conventional GCN-based models~\cite{mohamed2020social,wei2020view,luo2021robust,Park_2021_CVPR} that only focus on a single domain, our GCN-based AGFlow has the capability of \emph{learning to adapt} cross-domain information, fully incorporating scene information for comprehensive motion understanding.

% ------------------------------
\section{Methodology}
\label{sec:3}

\subsection{Problem Formulation}

Given a pair of input consecutive images, {\em i.e.}, source image ${\mathbf I}_1$ and target image ${\mathbf I}_2$, the task of optical flow estimation is to predict a dense displacement field between them. Deep learning based flow networks commonly employ an encoder-decoder pipeline to first extract context feature ${\mathbf f}_c$ and obtain motion cues ${\mathbf f}_m$, and then make flow prediction based on the fused feature ${\mathbf f}_o$ in a recurrent/coarse-to-fine manner.

In our approach, we represent the feature fusion in decoder as a graph-based reasoning and learning model, which is formulated as ${\mathbf f}_o = {\cal F}_{\\_{\cal G}}({\mathbf f}_c, {\mathbf f}_m)$. Specifically, we define the model as a directed graph $\cal G = (\cal{V, E})$, where $\cal V$ indicates a set of nodes, and $\cal E$ denotes edges specifying the connection and relation information among nodes. After $t$ runs of graph reasoning, the updated nodes are then mapped back to the original coordinate space to predict the displacement field.

\subsection{Adaptive Graph Reasoning for Optical Flow}

Fig.~\ref{fig:2} depicts an overview of the proposed adaptive graph reasoning for optical flow (AGFlow). Following the success of prior work~\cite{Jiang2021LearningTE}, we develop our AGFlow based on RAFT~\cite{Teed2020RAFTRA}. Specifically, given a pair of input images ${\mathbf I}_1$ and ${\mathbf I}_2$, we employ two residual block based encoders~\cite{He2016DeepRL} to extract a feature pair (${\mathbf f}_1$, ${\mathbf f}_2$) and context feature ${\mathbf f}_c$. Then, 4D correlation volumes are constructed on the feature pair in four scales. In the recurrent refinement framework, we utilize four convolutions to capture motion feature ${\mathrm f}_m$ from the multi-scale matching costs in each $9 \times 9$ region. After that, our adaptive graph reasoning (AGR) module takes the motion feature ${\mathrm f}_m$ and context feature ${\mathbf f}_c$ as inputs to perform a holistic motion reasoning. Please refer to RAFT~\cite{Teed2020RAFTRA} for more details about the implementations of the baseline model.

\paragraph{Node embedding.} 
The first step is to project context and motion features in regular coordinate space into graph space. The projection operation decouples positional information from the original grid feature, and makes the produced low dimensional node feature representations more compact and with sufficient expressive power. 
Here we divide the mapped nodes $\cal V$ in graph model into two groups: {\em context nodes} ${\mathbf v}_c = \{{v_{c}}^{\tt 1}, \cdots, {v_{c}}^{\tt n}\}$, containing appearance feature about shape and region information of scene context, and {\em motion nodes} ${\mathbf v}_m = \{{v_{m}}^{\tt 1}, \cdots, {v_{m}}^{\tt n}\}$, storing motion feature of cross-image matching dependency.

Specifically, given the context feature ${\mathbf f}_c \in \mathbb{R}^{c \times h \times w}$ and motion feature ${\mathbf f}_m \in \mathbb{R}^{c \times h \times w}$ from encoder network, we employ project function ${\cal P}_{{\mathrm f}\rightarrow {\mathrm v}}(\cdot)$ to assign features with similar representation to the same node. Let ${\mathbf v}_c \in \mathbb{R}^{C \times K}$ and ${\mathbf v}_m \in \mathbb{R}^{C \times K}$ denote the initial node embeddings in graph space, where $C$ indicates channel number and $K$ is the number of nodes. 

To build a global graph over a set of regions, we formulate ${\cal P}_{{\mathrm f}\rightarrow {\mathrm v}}(\cdot)$ as a linear combination of feature vector in grid space, {\em i.e.}, ${\mathbf v} = {\cal P}_{{\mathrm f}\rightarrow {\mathrm v}}({\mathbf f})$, and thus the produced nodes are able to aggregate long-range information within the overall original feature map. This is given by 

\begin{equation}
	{\mathrm v}_i = {\cal N}(\sum_{\forall j}{\cal F}_{{\mathrm f}\rightarrow {\mathrm v}}({\mathrm f})_{ij} \cdot {\mathrm f}_j),
	\label{eq1}
\end{equation} 
where ${\cal N}(\cdot)$ is a L-2 normalization function conducted on channel dimension of each node vector, and ${\cal F}_{{\mathrm f}\rightarrow {\mathrm v}}({\mathrm f}) \in \mathbb{R}^{N \times K}$ models the projection weights for mapping feature maps to node vectors.
% In practice, we simply implement ${\cal F}_{{\mathrm f}\rightarrow {\mathrm v}}(\cdot)$ with two convolution layers. 
Note that our approach can be trained with arbitrary input resolutions. In practice, we first use two convolutions on ${\mathbf f} \in \mathbb{R}^{c \times h \times w}$ to change the channel dimension from $c$ to $K$ so that a feature map with resolution of $K \times h \times w$ can be obtained. Then a reshape function is applied to produce ${\cal F}_{{\mathrm f}\rightarrow {\mathrm v}}({\mathrm f})$ with resolution $N \times K$, where $N = h \times w$, and K is a hyperparameter not relying on spatial resolution. Thus the two types of node embedding can be produced by ${\mathbf v}_c = {\cal P}_{{\mathrm f}\rightarrow {\mathrm v}}({\mathbf f}_c)$, and ${\mathbf v}_m = {\cal P}_{{\mathrm f}\rightarrow {\mathrm v}}({\mathbf f}_m)$.

\paragraph{Adaptive Graph Reasoning.}

Given node embeddings ${\mathbf v}$ in graph space, the adjacency matrix for graph reasoning can be commonly generated by measuring the similarity among all node vectors~\cite{Li2018BeyondGL}, as ${\mathbf A} = {\mathbf v}^{\mathsf T}{\mathbf v}$. After modeling the adjacency matrix ${\mathbf A}$, the graph reasoning with graph convolutional network~\cite{Kipf2017SemiSupervisedCW} is defined as
\begin{equation}
	\hat{\mathbf v} = {\cal F}_{\tt G}({\mathbf v}, {\mathbf A}) = \sigma({\mathbf A} {\mathbf v}^\mathsf{T} {\mathrm w}_{\tt G}),
\end{equation}
where $\sigma(\cdot)$ is a non-linear activation function, and ${\mathrm w}_{\tt G}$ is learnable parameters for graph convolution. $\hat{\mathbf v}$ is the updated node representations with graph reasoning, and it can be iteratively enhanced with more runs as $\hat{\mathbf v}^{(t)} = {\cal F}_{\tt G}({\mathbf v}, {\mathbf A})^{(t)}$, where $t$ denotes update iterations.

Let us consider the representation property of context and motion nodes. Motion nodes mainly encode the point-wise correspondence between an image pair while neglecting the intra-relations among pixels within regions, the context nodes, on the contrary, capture discriminative features for region and shape representations. Thus, we need to address two hurdles: First, 
there is an inevitable representation gap between context and motion nodes, which could hinder the effective information propagating of directly overall graph reasoning. Second, motion nodes lack constraints on shape or layout for the potential displacement field, and thus they are unable to yield enough context information for individual graph reasoning.

\begin{figure}[pt]
	\begin{center}
		\includegraphics[width=1\linewidth]{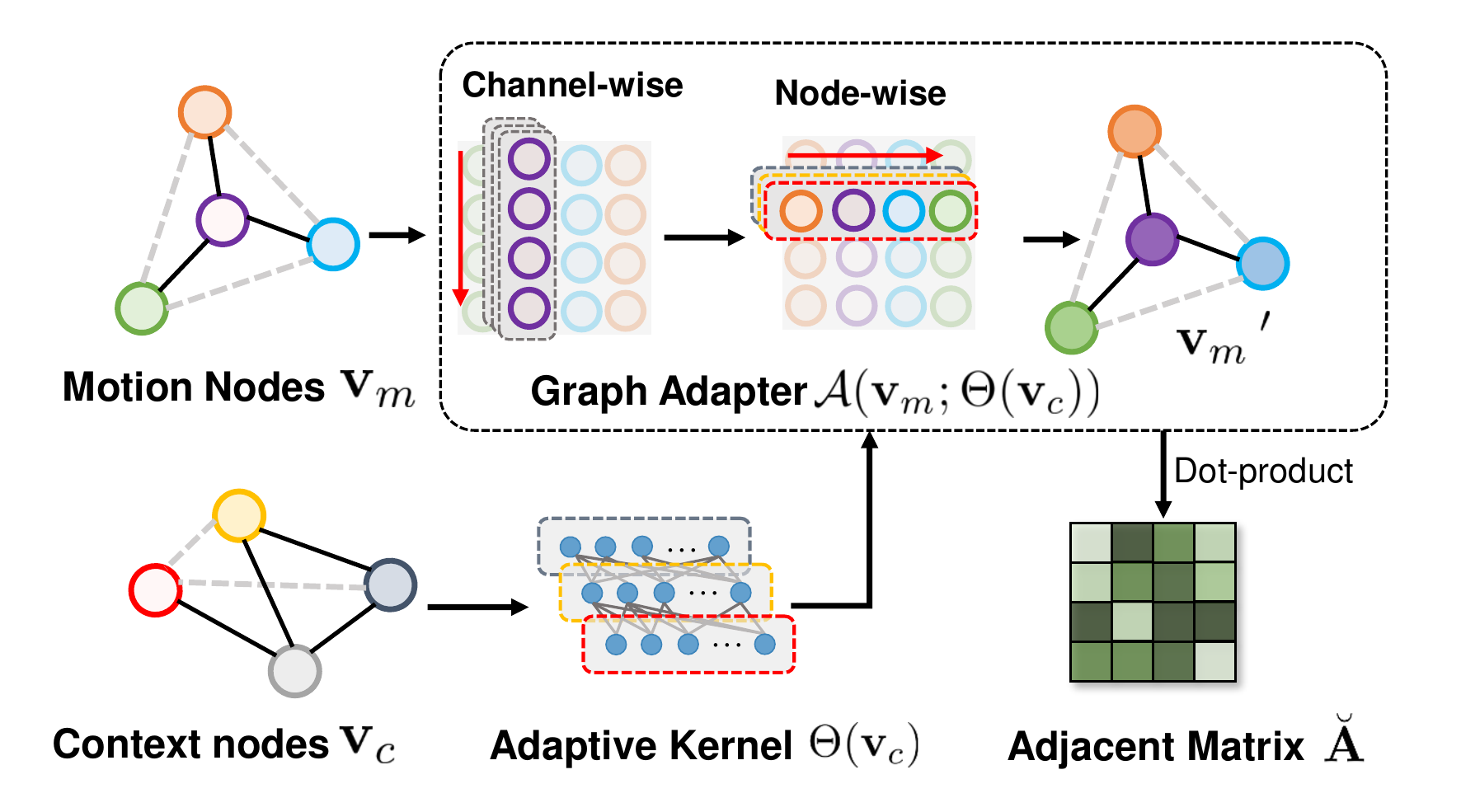}
	\end{center}
	\vspace{-1em}
	\caption{Architecture of Graph Adapter.}
	\vspace{-1em}
	\label{fig:3}
\end{figure}

To tackle the issues, we propose an {\em adaptive graph reasoning} (AGR) module to decouple the context reasoning from the matching procedure, and simultaneously transfer the region and shape prior of scene context to motion nodes in a one-shot manner. The key idea is to exploit the discriminative representations of shape and region in global context to guide the learning of motion adjacency matrix with adaptive parameters. Inspired by~\cite{Bertinetto2016LearningFO}, we devise an adaptive procedure of adjacency matrix learning to predict dynamic parameters that tailor the motion relation modeling based on image-specific contextual information. This is given by 
\begin{equation}
	\breve{\mathbf A} = {\cal A}({\mathbf v}_m ; \Theta({\mathbf v}_c)),
\end{equation}
where $\Theta(\cdot)$ is a parameter learner, and ${\cal A}(\cdot)$ denotes a context-to-motion graph adapter (GA) equipped with dynamic weights from $\Theta({\mathbf v}_c)$. In practice, we implement the $\Theta(\cdot)$ with a linear projection function with a softmax activation. As shown in Fig.~\ref{fig:3}, ${\cal A}(\cdot)$ is implemented with a two-layer MLP, where the first regular linear function followed by a ReLU activation is applied to perform channel-wise learning, and then the second linear function with adaptive kernel $\Theta({\mathbf v}_c)$ is utilized to perform node-wise interaction for relation learning with context-to-motion adaptation. Specifically, given context nodes ${\mathbf v}_c \in \mathbb{R}^{C \times K}$, we predict the adaptive kernel $\Theta({\mathbf v}_c) \in \mathbb{R}^{K \times K}$ with convolutions on channel dimension ($C \rightarrow K$). Then, we transfer it into the adaptive weights with shape $K \times K$ of the second linear function, which is used to produce ${{\mathbf v}_m}'$. Finally, we apply the dot-product similarity on ${{\mathbf v}_m}'$ to predict $\breve{\mathbf A}$.

The produced parameter $\Theta({\mathbf v}_c)$ relies on context nodes for dynamically leveraging the shape and region information of current input. Thus, motion nodes can be fast adaptive to scene context and favorably make full use of the transferred node relation for motion sub-graph reasoning. 
Thus the enhanced context nodes $\hat{\mathbf v}_c^{(t)}$ are produced by
\begin{equation}
	\hat{\mathbf v}_c^{(t)} = {\cal F}_{\tt G}({\mathbf v}_c, {\mathbf A})^{(t)}, {\tt where}~ {\mathbf A} = {\mathbf v}_c^{\mathsf T}{\mathbf v}_c,
\end{equation}
and similarly, motion nodes $\hat{\mathbf v}_m^{(t)}$ are produced by 
\begin{equation}
	\hat{\mathbf v}_m^{(t)} = {\cal F}_{\tt AG}({\mathbf v}_m, \breve{\mathbf A})^{(t)}, {\tt where}~ \breve{\mathbf A} = {\cal A}({\mathbf v}_m; \Theta({\mathbf v}_c)),
\end{equation} 
and ${\cal F}_{\tt AG}(\cdot)$ denotes motion nodes reasoning with adaptive graph convolutional network (AGCN).

\begin{table*}[ht]
	\centering
	\resizebox{0.92\textwidth}{!}{
		\begin{tabular}{clccccccc}
			\toprule
			\multirow{2}{*}{Training Data} & \multirow{2}{*}{Method} & \multicolumn{2}{c}{\underline{Sintel (val)}} &  \multicolumn{2}{c}{\underline{KITTI-15 (val)}} & \multicolumn{2}{c}{\underline{Sintel (test)}} & \multicolumn{1}{c}{\underline{KITTI-15 (test)}} \\
			& & Clean & Final & EPE & F1-all & Clean & Final & F1-all \\
			\midrule    
			\multirow{9}{*}{C + T} 
			& HD3\cite{Yin2019HierarchicalDD}            & 3.84  & 8.77 & 13.17 & 24.0 & - & - & - \\
			& LiteFlowNet\cite{Hui2018LiteFlowNetAL}    & 2.48  & 4.04  & 10.39 & 28.5 & - & - & - \\
			& PWC-Net\cite{Sun2018PWCNetCF}        & 2.55  & 3.93 & 10.35 & 33.7 & - & - & - \\
			%& LiteFlowNet2\cite{liteflownet2}   & 2.24  & 3.78  & 8.97 & 25.9 & - & - & - \\
			%& VCN\cite{vcn}            & 2.21  & 3.68  & 8.36 & 25.1 & - & -     & - \\ 
			%& MaskFlowNet\cite{maskflownet} & 2.25 & 3.61 & - & {23.1} & - & - & - \\ 
			& FlowNet2\cite{Ilg2017FlowNet2E}       & {2.02}  & \ 3.54 & 10.08 & 30.0 & 3.96  & 6.02 & - \\
			& DICL\cite{Wang2020DisplacementInvariantMC} & 1.94 & 3.77 & 8.70 & 23.6 & - & - & - \\
			& RAFT\cite{Teed2020RAFTRA}        & 1.43 & {2.71} & {5.04} & {17.4} & - & - & - \\
			& SCV\cite{Jiang2021LearningOF} & \bf{1.29} & 2.95 & 6.80 & 19.3 & - & - & - \\
			& \bf AGFlow (ours) & {1.31} & \bf{2.69} & \bf{4.82} & \bf{17.0} & - & - & - \\
			\midrule
			
			\multirow{9}{1.1cm}{C + T \\ + S + K \\ (+ H)} 
			%& FlowNet2 \cite{flownet2}  & (1.45) & (2.01) & (2.30) & (6.8) & 4.16  & 5.74 & 11.48  \\
			& PWC-Net+\cite{Sun2020ModelsMS}   & (1.71)     & (2.34)  & (1.50) & (5.3)  & 3.45  & 4.60 & 7.72 \\
			%& LiteFlowNet2 \cite{liteflownet2} & (1.30) & (1.62) & (1.47) & (4.8) & 3.48  & 4.69 & 7.74 \\
			%& HD3 \cite{hd3}         & (1.87)     & (1.17) & (1.31) & (4.1)  & 4.79  & 4.67 & 6.55 \\
			& IRR-PWC \cite{Hur2019IterativeRR}     & (1.92) & (2.51) & (1.63) & (5.3) & 3.84  & 4.58  & 7.65 \\
			& VCN \cite{Yang2019VolumetricCN}            & (1.66)     & (2.24) & (1.16) & (4.1) & 2.81  & 4.40 & 6.30 \\
			& MaskFlowNet\cite{Zhao2020MaskFlownetAF} & - & - & - & - & 2.52 & 4.17 & 6.10 \\
			& ScopeFlow\cite{BarHaim2020ScopeFlowDS} & - & - & - & - & 3.59 & 4.10 & 6.82 \\
			& DICL\cite{Wang2020DisplacementInvariantMC}  & (1.11) & (1.60) & (1.02) & (3.6) & 2.12 & 3.44  & 6.31 \\
			& RAFT\cite{Teed2020RAFTRA}  & {(0.76)} & {(1.22)} & {(0.63)} & {(1.5)} & {1.61}* & {2.86}* & {5.10} \\
			& SCV\cite{Jiang2021LearningOF} & (0.79) & (1.70) & (0.75) & (2.1) & 1.77* & 3.88* & 6.17 \\
			& \bf AGFlow (ours) & \bf{(0.65)} & \bf{(1.07)} & \bf{(0.58)} & \bf{(1.2)} & \bf 1.43* & \bf 2.47* & \bf 4.89 \\
			% \cmidrule[\lightrulewidth](r{0.3em}){2-9}
			% & RAFT-{\scriptsize WS}\cite{Teed2020RAFTRA}  & {(0.77)} & {(1.27)} & - & - & {1.61} & {2.86} & - \\
			% & SCV-{\scriptsize WS}\cite{Jiang2021LearningOF} & (0.86) & (1.75) & - & - & 1.77 & 3.88 & - \\
			% & \bf AGFlow-{\scriptsize WS} (ours) & \underline{(0.67)} & \underline{(1.12)} & - & - & \bf 1.43 & \bf 2.47 & - \\
			\bottomrule
		\end{tabular}
	}
	\caption{Quantitative comparison with state-of-the-art methods using EPE and F1-all metrics (the lower the better). Following previous works~\cite{Wang2020DisplacementInvariantMC, Teed2020RAFTRA, Jiang2021LearningOF}, we compare our results with all published works on three passes from two standard benchmarks. ``C + T'' indicates models are pretrained on FlyingChairs(C) and FlyingThing(T) to test generalization performance. ``+ S + K (+ H)'' denotes the training data combining Sintel(S), KITTI(K) and HD1K(H). ``+H'' with brackets means it is optional for some works~\cite{Teed2020RAFTRA, Hui2018LiteFlowNetAL}. ``*'' denotes the results with warm-start testing~\cite{Teed2020RAFTRA}. The best results are marked in {\bf bold} for better comparison.
	} \label{tab:1}
	\vspace{-1em}
\end{table*}

\paragraph{Attentive Readout.}

After $t$ runs of relation reasoning and state updating, we present an attentive readout module to project the enhanced context nodes ${\mathbf v}_c^{(t)}$ and motion nodes $\hat{\mathbf v}_m^{(t)}$ from graph space back to grid feature space, making the overall graph interaction model compatible with existing flow networks. Then the updated feature maps contain both global contextual information and local pixel-wise matching cost, which are properly used to make a better prediction for the flow field.

In particular, we formulate the reverse projection as
\begin{equation}
	\hat{\mathbf f} = {\cal P}_{{\mathrm v}\rightarrow {\mathrm f}}(\hat{\mathbf v}),
\end{equation}
where ${\cal P}_{{\mathrm v}\rightarrow {\mathrm f}}(\cdot)$ is a linear combination function that map the node vectors $\hat{\mathbf v} \in \mathbb{R}^{C \times K}$ to the feature maps $\hat{\mathbf f} \in \mathbb{R}^{C \times N}$ in the original grid space of flow network. In practice, we reuse the projection matrix in the node embedding procedure. The projection matrix contains pixel-to-node assignments and preserves the spatial details, which is crucial for recovering the resolution of feature maps. Besides, no additional parameters are involved by reusing the region assignments, which also helps to reduce computational overhead. 

The context feature $\hat{\mathbf f}_c$ is produced by a residual operation as
\begin{equation}
	\hat{\mathbf f}_c = {\mathbf f}_c + \alpha {\cal P}_{{\mathrm v}\rightarrow {\mathrm f}}({\mathbf v}_c),
\end{equation}
where $\alpha$ denotes a learnable parameter that is initialized as $0$ and gradually performs a weighted sum. Similarly, motion feature $\hat{\mathbf f}_m$ is produced by
\begin{equation}
	\hat{\mathbf f}_m = {\mathbf f}_m + \beta {\cal P}_{{\mathrm v}\rightarrow {\mathrm f}}({\mathbf v}_m).
\end{equation}

Given the enhanced feature $\hat{\mathbf f}_c$ and $\hat{\mathbf f}_m$, one potential hurdle for feature fusion is that context feature lacks correspondence information for cross-image matching, which could lead to a shift of global displacement and thus affect the flow accuracy. Therefore, we devise an attentive fusion function, which first learns to predict a set of scale weights from motion feature $\hat{\mathbf f}_m$ and then leverages them to implement global adjustment for entire dense displacement. Specifically, the attentive fusion function is defined as
\begin{equation}
	{\mathbf f}_o = (1+{\cal F}_{\tt CA}(\hat{\mathbf f}_m)) \hat{\mathbf f}_c \oplus \hat{\mathbf f}_m,
\end{equation}
where $\oplus$ is a concatenation operation, and ${\cal F}_{\tt CA}(\cdot)$ is a channel attention function~\cite{Hu2020SqueezeandExcitationN}, here implemented with two convolutions with ReLU and sigmoid activation.

\begin{figure*}[ht]
	\begin{center}
		\includegraphics[width=0.98\linewidth]{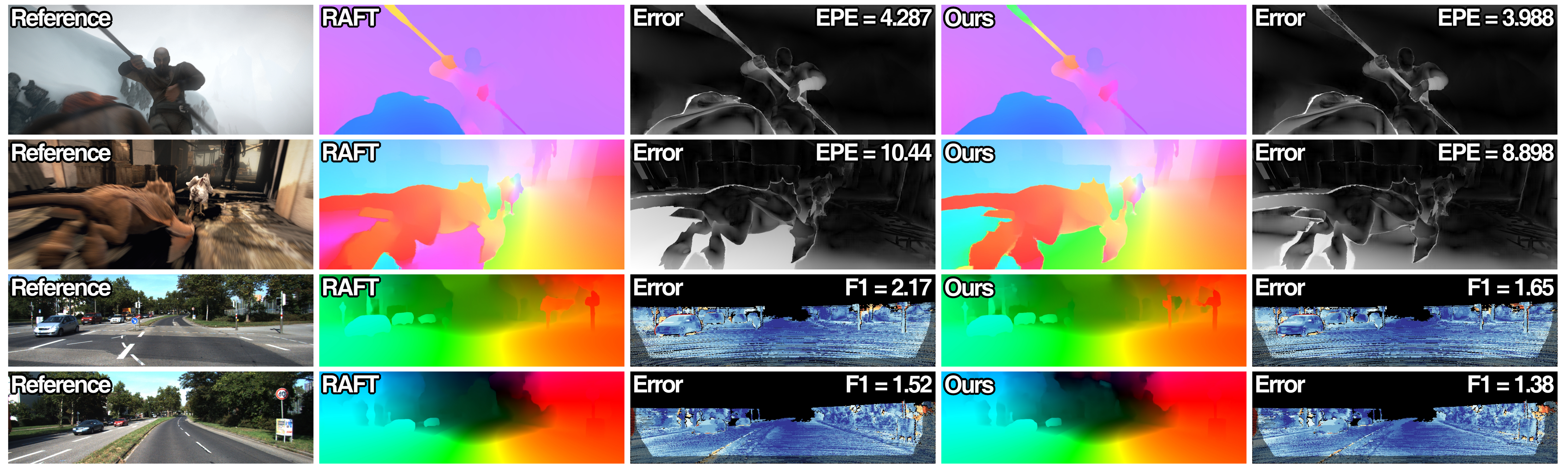}
	\end{center}
	\vspace{-1em}
	\caption{Qualitative comparisons with RAFT~\cite{Teed2020RAFTRA} on Sintel and KITTI {\tt test} set. All results are provided by the official website of each dataset. Best viewed in color.}
	\label{fig:both}
	\vspace{-1em}
\end{figure*}

\section{Experimental Results}

\subsection{Datasets and Evaluation Metrics}

We conduct extensive experiments on two standard datasets, {\em i.e.,} MPI-Sintel~\cite{Butler2012} and KITTI 2015~\cite{KITTI_2015}. We follow prior works~\cite{Teed2020RAFTRA, Jiang2021LearningOF} to utilize two standard evaluation metrics, {\em i.e.}, average end-point error (EPE) and the percentage of erroneous pixels $> 3$ pixels (F1-all), to evaluate the performance of predicted optical flow.

\subsection{Implementation Details}

The implementation of our approach is based on PyTorch toolbox. In our model, we set the number of context and motion nodes $K$ to $128$. The state updating iterations $t$ are set to $2$ and $1$ for context and motion graph, respectively. 

During training, we follow prior works~\cite{Teed2020RAFTRA, Jiang2021LearningTE} to adopt AdamW optimizer with one-cycle learning rate policy, and conduct model pretraining on synthetic data as the standard optical flow training procedure. The model is pretrained on FlyingChairs~\cite{Dosovitskiy2015FlowNetLO} for 180k iterations and then on FlyingThings~\cite{Mayer2016ALD} for 180k iterations. After that, we fine-tune the model on combined data from Sintel~\cite{Butler2012}, KITTI-2015~\cite{KITTI_2015}, and HD1K~\cite{Kondermann2016TheHB} for 180k iterations, and then submit the flow prediction to Sintel server for online evaluation. Finally, additional 50k iterations of finetuning are performed on KITTI-2015~\cite{KITTI_2015} for KITTI online evaluation. Our model is trained on 2 NVIDIA GeForce GTX 2080Ti GPUs, and the batch size is set to $8$ for better leveraging the GPU memory.

\subsection{Comparison with State-of-the-Arts}

\paragraph{Results on Sintel.}

On the training set of FlyingChairs (C) + FlyingThings (T), as shown in Tab.~\ref{tab:1}, our approach achieves an average EPE of $1.31$ on clean pass of Sintel dataset, which is competitive with SCV~\cite{Jiang2021LearningOF} and lower than the well-known RAFT~\cite{Teed2020RAFTRA} by $8.4\%$ ($1.43 \rightarrow 1.31$). On final pass, it obtains a score of $2.69$ in EPE, outperforming previous state-of-the-art methods SCV and RAFT by $8.8\%$ ($2.95 \rightarrow 2.69$) and $0.7\%$ ($2.71 \rightarrow 2.69$), respectively. The results demonstrate the good cross dataset generalization of our model.

On Sintel test set, we follow prior works~\cite{Hui2018LiteFlowNetAL, Hur2019IterativeRR, Teed2020RAFTRA} to submit the predicted flow to the official server for online evaluation. Our AGFlow achieves an EPE of $1.68$ on on Sintel clean pass, which surpasses top-ranked methods RAFT~\cite{Teed2020RAFTRA} and SCV~\cite{Jiang2021LearningOF} by $13.4\%$ ($1.94 \rightarrow 1.68$) and $2.3\%$ ($1.72 \rightarrow 1.68$), respectively. Besides, it obtains EPE = $2.83$ on final pass, outperforming recent SCV by a large margin ($21.4\%$). When utilizing the warm-start strategy, our approach sets new records of $1.43$ EPE on clean pass and $2.47$ on final pass, which significantly surpasses the previous best results by $11.2\%$ ($1.61 \rightarrow 1.43$) and $13.6\%$ ($2.86 \rightarrow 2.47$), respectively. It is worth noting that our AGFlow ranks 1st on final pass of Sintel benchmark among all approaches at the time of submission. 

Fig.~\ref{fig:both} (line 1 and 2) provides some qualitative comparisons with RAFT~\cite{Teed2020RAFTRA} on the challenging final pass of Sintel dataset, which demonstrates that our AGFlow is able to fully exploit scene context to effectively assist motion estimation with shape and region constraints, leading to achieving more accurate flow fields with clear motion boundaries.

\paragraph{Results on KITTI.}

We also provide the results of our approach on KITTI-15 dataset. As in Tab.~\ref{tab:1}, when training on C + T, our AGFlow achieves an average EPE of $4.82$ and F1-all score of $17.0\% $ on KITTI-15 validation set, which significantly surpass recent method SCV~\cite{Jiang2021LearningOF} by $29.1\%$ ($6.80 \rightarrow 4.82$) and $11.9\%$ ($19.3 \rightarrow 17.0$), respectively. For online evaluation, our approach achieve new state-of-the-art performance of $4.89\%$ in F1-all, outperforming top-ranked methods SCV~\cite{Jiang2021LearningOF} and RAFT~\cite{Teed2020RAFTRA} by $20.7\%$ ($6.17\rightarrow 4.89$) and $4.1\%$ ($5.10 \rightarrow 4.89$), respectively. Some qualitative comparisons with RAFT~\cite{Teed2020RAFTRA} on KITTI dataset are illustrated in Fig.~\ref{fig:both} (line 3 and 4), which shows that the proposed global reasoning conditioned on scene context helps to decrease the uncertainty of ambiguous matching in some tough regions.

% ----------
\subsection{Ablation Analysis} \label{sec:4}

\paragraph{Comparison with Grid Feature Enhancement.}
% Conventional Feature Fusion {\em vs.} Graph-based Reasoning
We first compare the proposed AGFlow with widely-used methods for optical flow that enhance features in regular grid space, including RAFT~\cite{Teed2020RAFTRA}, dense and dilated convolutions~\cite{Sun2018PWCNetCF, Hur2019IterativeRR}. As shown in Tab.~\ref{tab:cmpr}, dense and dilated convolutions slightly improve the flow accuracy with heavy model complexity. In contrast, our AGFlow achieves better performance, yet only needs additional $0.30$ M parameters, reducing parameters by around $90\%$. This demonstrates that the proposed low dimensional graph reasoning scheme is effective to boost the flow accuracy in an efficient manner.

\paragraph{Comparison with SuperGlue~\cite{Sarlin2020SuperGlueLF}.} SuperGlue is a well-known method that employs graph neural network for feature matching. As can be seen in Tab.~\ref{tab:cmpr} (line 4), the original SuperGlue on grid feature requires a large amount of GPU memory, which is out of range with general settings for model training. Thus we re-implement SuperGlue with our low dimensional graph model (termed G-SuperGlue). Compared with G-SuperGlue, our AGFlow not only achieves considerable performance gain ($8.0\%$), but also reduces parameters by $85\%$. This is because our adaptive graph reasoning scheme is simple yet effective, and capable of fast transferring the region and shape prior from scene context to motion nodes in a one-shot manner.

\paragraph{Comparison with GCU~\cite{Li2018BeyondGL}.}
We also compare our AGFlow with the basic region-based graph reasoning model~\cite{Li2018BeyondGL}. Since GCU projects all feature maps into a single type of node, it requires fewer parameters than other methods. However, as we mention above, the representation gap between context and motion feature hinders the effectiveness of relation reasoning with a simple graph, thus resulting in a performance drop on both passes of Sintel. We regard it as the base graph model, as in line 5 of Tab.~\ref{tab:cmpr}. In contrast, we carefully project feature maps into context and motion nodes, and further propose an adaptive graph reasoning approach to perform the {\em task-specific} hybrid reasoning, allowing our model to significantly reduce the average end-point error by around $14.8\%$.

\begin{table}[t]
	\centering
	\resizebox{0.46\textwidth}{!}{
		\begin{tabular}{clcccccc}
			\toprule
			\multirow{2}{1cm}{\centering} & \multirow{2}{*}{\centering Method} & \multirow{2}{0.1cm}{ } & \multirow{2}{1.8cm}{\centering Param (M) $\downarrow$} & \multicolumn{2}{c}{\centering Sintel (val) EPE $\downarrow$} \\
			\cmidrule(lr){5-6}
			\cmidrule(lr){7-8}
			& & & & Clean & Final \\
			\midrule
			\multirow{4}{1cm}{\centering Grid Feature} & ~~ RAFT &   & 5.26  & 1.65 & 3.04 \\
			& + Dense Convs &  & + 3.34 & 1.63 & 2.98 \\
			& + Dilated Convs &  & + 0.85 & 1.66 & 3.02 \\
			& + SuperGlue &  & + 2.99 & - & - \\  % ~\cite{Sarlin2020SuperGlueLF}
			\midrule
			\multirow{3}{1cm}{\centering Graph Space} & ~~ GCU (Base) &  & 5.44 & 1.76 & 3.14 \\
			& + G-SuperGlue &  & + 1.82 & 1.63 & 3.01 \\ 
%			& + GCU &  & + 0.04 & 1.79 & 3.17 \\  % ~\cite{Li2018BeyondGL}
			& + \bf AGR (ours)  &   & + 0.12 & \bf 1.50 & \bf 2.88 \\			
			\bottomrule
		\end{tabular}
	}
	\caption{Quantitative comparisons with related methods (refer to Sec.~\ref{sec:4} for more details). We set RAFT~\cite{Teed2020RAFTRA} as the baseline model for the method with regular gird space enhancement. The methods in each part are plugged into the same baseline and trained on C + T ($180k$) for fair comparison.}
	\label{tab:cmpr}
	\vspace{-.5em}
\end{table}

\begin{table}[t]
	\centering
	\resizebox{0.46\textwidth}{!}{
		\begin{tabular}{cccccc}
			\toprule
			\multirow{2}{*}{\centering } & \multirow{2}{*}{\centering Settings} & \multirow{2}{1.1cm}{\centering FLOPs (G)} & \multirow{2}{1.1cm}{\centering Param (M)} & \multicolumn{2}{c}{\centering Sintel (val) EPE} \\
			\cmidrule(lr){5-6}
			& & & & Clean & Final \\
			\midrule
			\multirow{4}{1.5cm}{\centering Graph Reasoning } & Base Graph & 381.06 & 5.44 & 1.76 & 3.14 \\
			& + SGR (no GA) & + 16.95 & + 0.07 & 1.65 & 2.97  \\
			& \underline{+ AGR} & + 17.09 & + 0.12 & 1.50 & 2.88 \\
			\midrule
			\multirow{4}{1.2cm}{\centering Node numbers} 
			& $K=32$ & + 3.09 & + 0.04 & 1.61 & 3.02 \\
			& $K=64$ & + 5.88 & + 0.07 & 1.56 & 2.95 \\ 
			& \underline{$K=128$} & + 17.09 & + 0.12 & 1.50 & 2.88 \\
			& $K=256$ & + 61.84 & + 0.2 & 1.49 & 2.90 \\  
			\midrule
			\multirow{2}{1.2cm}{\centering Attentive readout}      
			& \underline{On}  & + 17.09 & + 0.12 & 1.50 & 2.88 \\
			& Off & + 16.82 & + 0.10 & 1.55 & 2.94 \\
			\bottomrule
		\end{tabular}
	}
	\caption{Ablation analysis for different settings of our AGFlow. ``SGR'' indicates separated graph reasoning for context and motion nodes ({\em i.e.}, without graph adapter), and ``AGR'' denotes overall adaptive graph reasoning. All methods are trained on C + T ($180k$) for fair comparison. \underline{Underline} indicates the default settings in our model.}
	\label{tab:abla}
	\vspace{-1em}
\end{table}

\begin{table}[t]
	\centering
	\resizebox{0.32\textwidth}{!}{
		\begin{tabular}{cccc}
			\toprule
			Method  &  & Param (M) $\downarrow$ & Time (ms) $\downarrow$\\
			\midrule
			RAFT &  & 5.26 (-) & 86.9 (-)  \\
			GMA &  & 5.8 (+ 0.54) & 113.8 (+ 26.9) \\
			\bf AGFlow  &  & 5.56 (+ 0.30) & 90.7 (+ 3.8) \\			
			\bottomrule
		\end{tabular}
	}
	\vspace{-.5em}
	\caption{Computational comparisons with state-of-the-arts on a single Geforce RTX 2080Ti GPU.}
	\label{tab:time}
	\vspace{-.5em}
\end{table}

\paragraph{Effectiveness of Adaptive Graph Reasoning.}
In Tab.~\ref{tab:abla}, we empirically analyze the computational cost and corresponding performance gain of the core component in the proposed AGFlow. Using separated context and motion reasoning based on graph model boosts the performance by $5.4\% \sim 6.3\%$. Besides, we further incorporate the proposed graph adapter into the context-to-motion interaction, and then yield the proposed adaptive graph reasoning (AGR) module, which brings about $9\%$ additional performance gain with only $0.14$ G FLOPs and $0.05$ M parameters for extra computational overhead.

\paragraph{Ablation for Node Numbers.}
We empirically show the influences of node numbers $K$ in our graph model. As shown in Tab.~\ref{tab:abla}, when more nodes are used ($32 \rightarrow 64 \rightarrow 128$), the average end-point error are gradually decreased from $1.61$ on Sintel clean pass and $3.02$ on final pass to $1.50$ and $2.88$, respectively. However, if furthermore nodes are involved ($128 \rightarrow 256$), the flow accuracy almost remains the same and the computational overhead is largely increased by $2.6$ times. This is because some redundant feature representations are generated with nodes, which brings no benefit to flow estimation. Therefore, we set $K = 128$ to ensure a good balance between efficiency and performance.

\paragraph{Ablation for Attentive Readout.}
We also test the influence of attentive readout compared with regular readouts~\cite{Li2018BeyondGL} in Tab.~\ref{tab:abla}. As can be seen, incorporating it into our model brings about $3\%$ in performance gain and only requires negligible computation cost and parameters, demonstrating the cost-effective property of this component. 

\paragraph{Runtime Comparison.}
We provide the parameters and runtime of state-of-the-art methods in Tab.~\ref{tab:time}. Compared with GMA~\cite{Jiang2021LearningTE}, our AGFlow can achieve competitive performance while reducing $0.24$ M parameters. Besides, the inference speed is boosted by $20.3\%$ ($113.8 \rightarrow 90.7$). The comparisons clearly demonstrate the effectiveness of our AGFlow.

% ------------------------------
\section{Conclusion}

In this paper, we present a novel graph-based approach termed adaptive graph reasoning for optical flow (AGFlow), which performs global reasoning to explicitly emphasize scene context and motion dependencies for flow estimation. The key idea is adaptive graph reasoning, which intends to fast enhance the feature representation of motion nodes conditioned on the global context with shape and boundary. Comprehensive experiments demonstrate that our AGFlow is effective and flexible to alleviate the matching ambiguities in challenging scenes, and sets new records in two standard flow benchmarks. We hope our work will offer a fresh perspective in re-thinking the design of optical flow models.

\paragraph{Acknowledgements.}
This work was supported by the National Key R\&D Plan of the Ministry of Science and Technology No. 2020AAA0104400, and the National Natural Science Foundation of China (NSFC) No.61872067 and No.61720106004.

\bibliography{aaai22}

\end{document}